\pdfoutput=1

\documentclass[11pt]{article}

\usepackage[final]{acl}

\usepackage{times}
\usepackage{latexsym}
\usepackage{booktabs}
\usepackage[T1]{fontenc}

\usepackage[utf8]{inputenc}
\usepackage{amssymb}
\usepackage{microtype}
\usepackage{multirow}
\usepackage{inconsolata}

\usepackage{graphicx}
\usepackage{flushend}

\title{Overview of the MEDIQA-OE 2025 Shared Task on Medical Order Extraction from Doctor-Patient Consultations}

\author{
 \textbf{Jean-Philippe Corbeil},
 \textbf{Asma Ben Abacha},
 \textbf{Jérôme Tremblay},
 \textbf{Phillip Swazinna},\\
 \textbf{Akila Jeeson Daniel},
  \textbf{Miguel Del-Agua},
 \textbf{Francois Beaulieu}
\\
 Microsoft Healthcare \& Life Sciences
\\
 \small{
   \textbf{Correspondence:} \href{mailto:jcorbeil@microsoft.com}{\{jcorbeil,abenabacha\}@microsoft.com}
 }}

\begin{document}
\maketitle
\begin{abstract}
Clinical documentation increasingly uses automatic speech recognition and summarization, yet converting conversations into actionable medical orders for Electronic Health Records remains unexplored. A solution to this problem can significantly reduce the documentation burden of clinicians and directly impact downstream patient care. We introduce the MEDIQA-OE 2025 shared task, the first challenge on extracting medical orders from doctor-patient conversations. Six teams participated in the shared task and experimented with a broad range of approaches, and both closed- and open-weight large language models (LLMs). In this paper, we describe the MEDIQA-OE task, dataset, final leaderboard ranking, and participants’ solutions.
\end{abstract}

\begin{table*}[t!]
    \centering
    \caption{Final Ranking of MEDIQA-OE competition on the test set (100 samples) in which 6 teams participated. Our two baselines (excluded from the ranking) are a simple one-shot prompt for MediPhi-Instruct (1) and GPT-4o-0806 (2) with one example from the training set.}
    \begin{tabular}{c|l|l|c|ccccc}
        \hline
         \textbf{\small Rank} & \textbf{\small Team Name} & \textbf{\small Method} & \textbf{\small Match} & \textbf{\small Desc.} & \textbf{\small Reason} & \textbf{\small Type} & \textbf{\small Prov.} & \textbf{\small AVG} \\
         \hline
         1 & WangLab & \multirow{2}{*}{\small \shortstack[l]{ GPT-4 constrained dec.\\ Detailed instructions}} & \textbf{81.8} & \textbf{66.8} & 29.5 & \textbf{81.5} & \textbf{63.0} & \textbf{60.2} \\
         & & & & & & & & \\
         2 & silver\_shaw & \multirow{2}{*}{\small \shortstack[l]{ Gemini 2.5 Pro \tiny{w/ thinking} \\ Detailed plan \& instructions}} & 76.4 & 64.1 & \textbf{41.3} & 74.7 & 60.4 & 60.1 \\
         & & & & & & & & \\
         3 & MISo KeaneBeanz & \multirow{2}{*}{\small \shortstack[l]{ Qwen3 32B Q4\_K\_M \tiny{w/o thinking} \\ Instructions w/ 2 shots}} & 73.4 & 58.0 & 35.6 & 71.6 & 48.4 & 53.4 \\
         & & & & & & & & \\
         4 & \small{EXL Health AI Lab} & \multirow{2}{*}{\small \shortstack[l]{ MedGemma 27B \\ One shot (short format) }} & 67.7 & 54.5 & 30.5 & 66.2 & 52.5 & 50.9 \\
         & & & & & & & & \\
         5 & MasonNLP & \multirow{2}{*}{\small \shortstack[l]{ Llama4 17B 16E Instruct \\ One shot w/o orders }} & 55.5 & 39.1 & 19.8 & 50.9 & 41.3 & 37.8 \\
         & & & & & & & & \\
         - & \small{\texttt{Baseline 2}} & \multirow{2}{*}{\small \shortstack[l]{ GPT-4o \\ Simple prompt w/ one shot }} & 63.6 & 39.5 & 20.4 & 59.3 & 1.0 & 30.1 \\
         & & & & & & & & \\
         - & \small{\texttt{Baseline 1}} & \multirow{2}{*}{\small \shortstack[l]{ MediPhi-Instruct 3.8B \\ Simple prompt w/ one shot }} & 43.3 & 25.8 & 19.5 & 39.6 & 13.8 & 24.7 \\
         & & & & & & & & \\
         6 & HerTrials &  \multirow{2}{*}{\small \shortstack[l]{ Llama3.2 3.2B \\ Instructions w/ one shot }} & 31.2 & 19.6 & 9.0 & 29.6 & 5.6 & 15.9 \\
         & & & & & & & & \\
         \hline
    \end{tabular}
    \label{tab:ranking}
\end{table*}

\section{Introduction}

In recent years, the burden of clinical documentation has reduced the time clinicians can devote to direct patient care, and ultimately limited the number of patients physicians can help. To mitigate this, many hospitals and clinics now deploy automatic speech recognition and note summarization tools during consultations. A natural next step in this pipeline is medical order extraction (e.g., medications, labs, imaging, follow-ups) from conversation transcripts to directly populate Electronic Health Records (EHRs).

\begin{figure}[!ht]
\centering
\includegraphics[width=\linewidth,trim={0.8cm 0.8cm 0.7cm 1cm},clip]{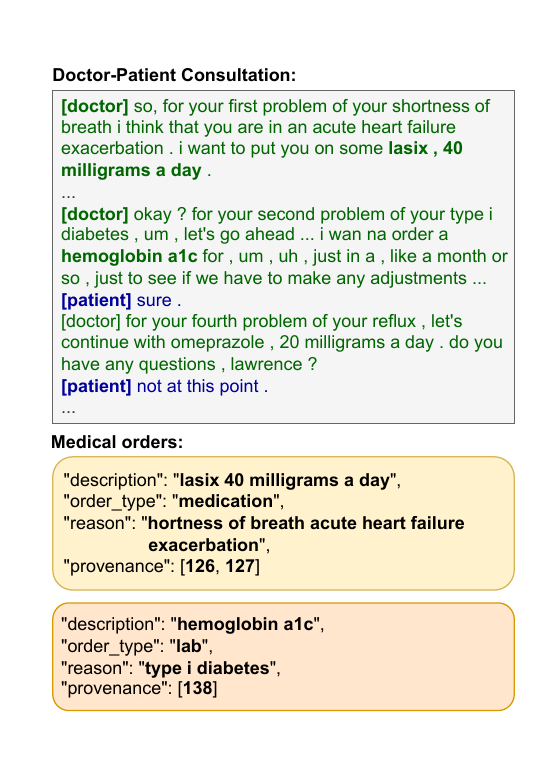}
\caption{The medical order extraction task takes a doctor-patient dialog and extracts a JSON list of orders containing four keys (description, order\_type, reason, and provenance). Orders that were previously prescribed but not explicitly renewed should be excluded (e.g. omeprazole in this example).}
\label{fig:oe_task}
\end{figure}

While named entity recognition (NER) and relation extraction (RE) have been extensively studied in clinical NLP\footnote{Natural Language Processing.} \cite{medex,doan2010,yang2020clinical,fabacher2025,henry20202018,lybarger2023n2c2}, extracting actionable, structured orders from full-length consultations remains underexplored despite its potential impact. The task is challenging: inputs are long, dialogues contain interruptions as well as revisions, and outputs combine schema-constrained fields (e.g., order type) with free-text attributes (e.g., description, reason). These challenges are compounded by distributional shifts, as clinicians adapt their language to patients without medical training during consultations.

In this era of LLMs \cite{brownlanguagegpt3,achiam2023gpt4}, new approaches have become feasible for the medical order extraction task --- combining long-context reasoning with schema-aware generation --- yet limitations in context length, controllability, and calibration persist. The MEDIQA-OE shared task\footnote{\url{https://sites.google.com/view/mediqa-2025}} investigates these challenges and benchmarks solutions to improve EHR clinical documentation, which we believe can reduce the burden on providers while ensuring the accurate capture of critical patient orders.

\section{Previous Work}

Tasks similar to order extraction in clinical NLP are commonly formulated as NER and RE. Early systems were rule-based (e.g., MedEx by \citet{medex}) or used classical machine learning such as support vector machines \cite{doan2010}. With pretrained contextual encoders, fine-tuned transformer models (e.g., BERT \cite{bert}, ClinicalBERT \cite{alsentzer-etal-2019-publicly}) became the standard for NER/RE and delivered consistent gains on clinical benchmarks \cite{yang2020clinical,fabacher2025}.

More recently, LLMs enable span-free formulations that cast extraction as reading-comprehension style generation. Prompting methods \cite{peng2023clinical, cui2023medtem2,PENG2024104630} have shown strong results on several clinical information extraction tasks, including adverse drug events \cite{henry20202018} and social determinants of health \cite{lybarger2023n2c2}. However, order extraction from full patient–doctor dialogues remains underexplored, particularly when models must (i) handle long, multi-speaker inputs and (ii) generate outputs that mix schema-constrained fields (e.g., order type, provenance) with free-text attributes (e.g., description, reason).

\section{Methodology}

\subsection{Source Datasets}

The long-form doctor-patient conversations used for the order-extraction task are primarily drawn from two datasets: ACI-Bench \cite{acibench} and PriMock57 \cite{primock57}. The ACI-Bench corpus comprises 207 naturalistic conversations between physicians and patients, curated by domain experts to reflect real-world clinical interactions. Similarly, the PriMock57 dataset contains 57 mock doctor-patient dialogues, designed to simulate clinical scenarios in a controlled setting.
Recent works such as Notechat \cite{wang2023notechat} has introduced large-scale synthetic dialogue datasets. While this corpus is the largest, we excluded it due to the prevalence of low-quality dialogues we observed.

\subsection{Annotations}

We asked medically trained annotators to produce the gold-standard medical orders for the high-quality conversations of Primock57 and ACI-Bench. Annotation guidelines instructed to assess every medical order of type medication, imaging, lab, or follow-up within the conversation the way a doctor would create them in the EHR. This was intended to replicate doctors' current process executed at the end of a patient encounter. We measured an inter-annotator agreement of $0.768$. We sampled 100 examples containing 255 medical orders across both data sources as a test set and kept the others as training set (64 samples) used for few-shot prompting, and development set (100 samples)~\cite{Corbeil-2025}.

\subsection{Evaluation}

We evaluate model performance across four key metrics: description, reason, type, and provenance. Results are reported after performing a matching between reference and hypothesis orders based on description field's word overlap\footnote{Necessary to compare orders with each other.}. An intermediary metric, the match score, is computed from this alignment process as the F1 between reference and predicted orders without looking at the content, thus specifically accounting for the amount of fabricated or omitted orders. It represents an upper bound for other metrics that are penalized for empty values for fair comparison. For description and reason metrics, we compute F1 scores of the rouge metric \cite{lin-2004-rouge} over unigrams. Type is evaluated using accuracy due to its limited label space, and provenance is assessed via F1 score over provenance labels\footnote{Turn numbers where the order originates in the transcript.}. Finally, the leaderboard ranking is assessed via the average of all four key metrics: description, reason, type, and provenance.

\section{Results}

\subsection{Leaderboard Ranking}

We provided in Table \ref{tab:ranking} the final leaderboard of the MEDIQA-OE along participants' approaches and our two baselines, which were used as reference points while being excluded from the ranking. All solutions are based on prompting language models. While there are two closed-source LLMs at the top of the ranking, the remaining submissions are leveraging open-weight LLMs in few-shot settings. WangLab obtained the $1^{st}$ rank of the competition by prompting GPT-4 \cite{achiam2023gpt4} with JSON-constrained decoding and detailed instructions. Following closely by 0.1\% on the average score, silver\_shaw \cite{silverShaw} achieved the $2^{nd}$ place by using Gemini 2.5 Pro \cite{comanici2025gemini} in thinking mode. The other approaches \cite{EXL,MasonNLP} leveraged different open-weight models in few-shot settings such as Qwen3 32B \cite{qwen3_think_deeper_act_faster_2025}, MedGemma 27B \cite{sellergren2025medgemma}, Llama4 Scout 17B \cite{meta_llama4_multimodal_2025} and Llama3.2 3.2B \cite{meta_llama3_2_edge_vision_2024}. Participants only appended one or two shot(s) examples due context limitations from long input-output pairs, and some even reduced examples into shorter formats. Overall, they also wrote simpler prompts compared to the two closed-weight LLM solutions.

\subsubsection{WangLab's Approach}

WangLab won the competition by prompting GPT-4 \cite{achiam2023gpt4} in a zero-shot setting. They obtain an average score of 60.2\% with JSON-constrained decoding on the order format as well as using very detailed rules in the instructions. They achieved the highest match score at 81.8\%, which indicates that a large proportion of reference orders are well matched. The average gains are double digits over the \texttt{Baseline 1} based on GPT-4o with improvements on the provenance (+62.0\%), description (+27.3\%), type (+22.2\%) and reason (+9.1\%) scores.

Their prompt provides very detailed instructions, and is as follows: 
\begin{enumerate}
    \item Role attribution
    \item Transcript definition with example
    \item Task definition
    \item Type definitions with rules and examples
    \item Output JSON key definitions
    \item Reason guidelines with specific examples
    \item JSON output example
    \item Overall guidelines
    \item Eliciting JSON output
\end{enumerate}

\subsubsection{silver\_shaw's Approach}

 Following closely 0. 1\% on the average score, silver\_shaw \cite{silverShaw} achieved $2^{nd}$ position with the highest reason score at 41. 3\% by prompting Gemini 2.5 Pro \cite{comanici2025gemini} in thinking mode with a detailed reasoning plan and instructions.
 
 Their one-call prompting approach asks the model to proceed in three steps aimed at mirroring the clinical reasoning processes: chain-of-thought analysis, self-critique \& verification, and deterministic JSON generation.

\subsubsection{MISo KeaneBeanz's Approach}

MISo KeaneBeanz's approach reached the $3^{rd}$ rank by prompting the 4-bit quantized open-weight model Gwen3 32B \cite{qwen3_think_deeper_act_faster_2025} in a two-shot setting.

\subsubsection{EXL Health AI Lab's Approach}

EXL Health AI Lab achieved the $4^{th}$ rank at 50.9\% leveraging a one-shot solution prompting MedGemma 27B \cite{sellergren2025medgemma}, an open-weight medical LLM. Their experiments covered agentic workflows such as ReAct \cite{yao2023react} and a four-step multi-agent pipeline. The one-shot method remained more accurate potentially because of the negative impact of noises introduced by multi-step approaches.

\subsubsection{MasonNLP's Approach}

The $5^{th}$ rank of the MEDIQA-OE competition was attributed to MasonNLP \cite{MasonNLP}. They used Llama4 17B \cite{meta_llama4_multimodal_2025} in a minimal one-shot prompting setting. The authors also reported an experiment with Llama4 8B.

\subsubsection{HerTrials' Approach}

HerTrials team ranked $6^{th}$ with a one-shot prompting of the smallest open-weight language models Llama3.2 \cite{meta_llama3_2_edge_vision_2024}.

\subsection{Analysis of Open-weight LLMs}

\begin{figure}[ht!]
\centering
\includegraphics[width=\linewidth]{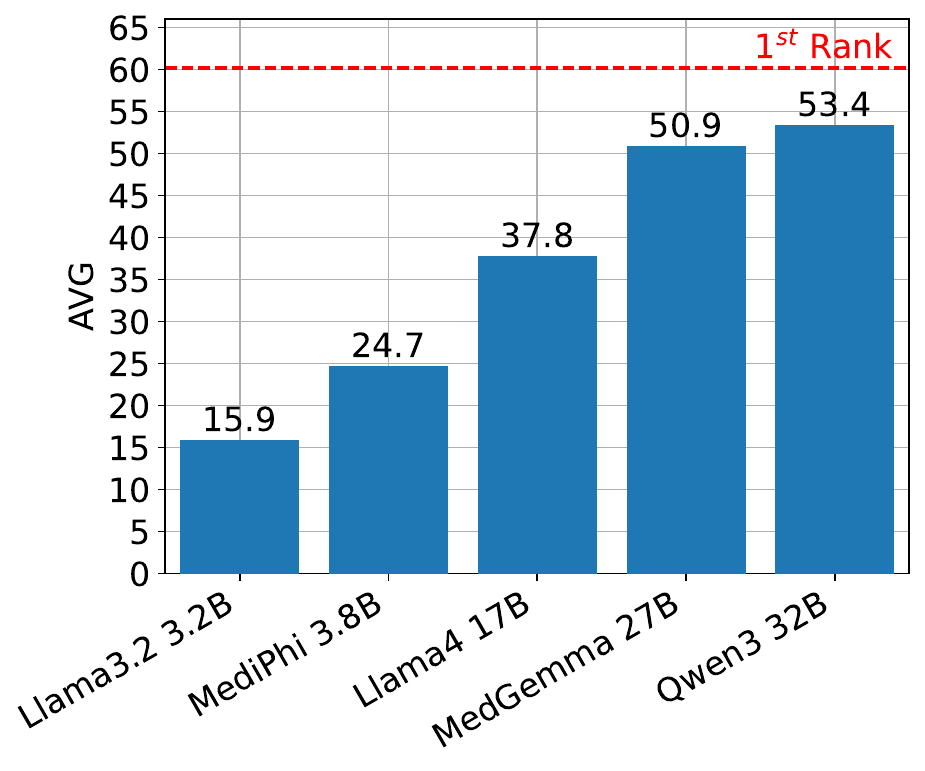}
\caption{Open-weight models ranking obtained with few shots correlates with parameter count.}
\label{fig:oss_models}
\end{figure}

We show the correlation between final accuracy and open-weight model sizes in Figure \ref{fig:oss_models}. Despite prompt variations, we computed a strong Pearson correlation of 0.981 between leaderboard ranking and model sizes, which is in line with previous work in clinical NLP \cite{dada-etal-2025-biomedical}.

\section{Discussion and Limitations}

In spite of top-ranking solutions achieving considerable scores with zero- and few-shot prompting and reasoning, significant gaps remain to push further the performance of the medical order-extraction task.\\
\textbf{First}, we notice that the maximum match F1 score is of 81.8\%, which means that there is still room of nearly 20\% to match the number of orders in the reference.\\
\textbf{Second}, description and provenance are both lagging behind the raw match score (i.e., their upperbound) by approximately 15-18\%. Provenance was considerably improved in this challenge by using: larger models, constrained decoding, and specific instructions. Future work could explore embedding-based and hybrid systems.\\
\textbf{Third}, order types are all very close to match scores, which highlights how such classification tasks are well suited for LLMs.\\
\textbf{Fourth}, we observe low performances on the reason field which might come from the dispersion of reasons across the conversation and the fact that it is an optional field with scarcer annotations.

One of the main limitations of the current task is the small dataset sizes. The current trainset size of 64 samples limits the ability to use it for finetuning --- which could particularly make open-weight small language models more competitive. Future work might produce larger datasets or leverage synthetic ones. While the inter-annotator agreement is considerably high, annotations might also present noises (e.g., span boundaries, non-expert conversational style instead of formal writing, etc.) which limit the maximum score below 100\%.

\section{Conclusion}

To conclude, the medical order-extraction task was tackled by a variety of zero- and few-shot approaches using open- and closed-weight LLMs. Closed-weight models such as GPT-4 and Gemini 2.5 Pro in zero-shot setting dominated the top ranks, leveraging detailed instructions, constrained decoding and reasoning. We observed a significant correlation of 0.981 between open-weight model sizes in few-shot settings and final accuracy. Although final scores considerably improved over the baselines especially in the match and provenance metrics, we still observe a significant gap in total extracted orders performance from the match score of 81.8\% as well as in performances on the description and the reason free-form fields. We believe future works include synthetic data generation, model fine-tuning, hybrid systems, and focus on improving small language models.


\bibliography{custom}




\end{document}